\documentclass[preprint,review,12pt]{elsarticle}

\usepackage{amssymb}
\usepackage{algorithmicx,algorithm}
\usepackage{hyperref}
\usepackage{multirow}
\usepackage{stfloats}
\usepackage{graphicx}
\usepackage{color}
\usepackage[justification=centering]{caption}
\usepackage{url}
\usepackage{xcolor}
\usepackage{hyperref}

\usepackage{textcomp}

\usepackage{caption}
\usepackage{diagbox}
\usepackage{float}
\usepackage{bm}
\usepackage{booktabs}
\usepackage{float}
\usepackage{multirow}
\usepackage{soul}
\soulregister\cite7 
\soulregister\citep7 
\soulregister\citet7 
\soulregister\ref7 
\soulregister\pageref7 
\usepackage{appendix}

\begin{document}


\begin{frontmatter}

\title{CVM-Cervix: A Hybrid Cervical Pap-Smear Image Classification Framework Using CNN, Visual Transformer and Multilayer Perceptron}

\author[a]{Wanli Liu}

\author[a]{Chen Li\corref{mycorrespondingauthor}}
\cortext[mycorrespondingauthor]{Corresponding author}
\ead{lichen201096@hotmail.com}

\author[b]{Ning Xu}

\author[c]{Tao Jiang}

\author[a]{Md Mamunur Rahaman}

\author[d]{Hongzan Sun}

\author[e]{Xiangchen Wu}

\author[a]{Weiming Hu}

\author[a]{Haoyuan Chen}

\author[a,f]{Changhao Sun}

\author[g]{Yudong Yao}

\author[h]{Marcin Grzegorzek}

\address[a]{Microscopic Image and Medical Image Analysis Group, College of Medicine and Biological Information Engineering, Northeastern University, Shenyang, China}

\address[b]{Liaoning Petrochemical University, Fushun, China}

\address[c]{School of Control Engineering, Chengdu University of Information Technology, Chengdu, China}

\address[d]{Shengjing Hospital, China Medical University, Shenyang, China}

\address[e]{Suzhou Ruiqian Technology Company Ltd., Suzhou, China}

\address[f]{Shenyang Institute of Automation, Chinese Academy of Sciences, Shenyang, China}

\address[g]{Department of Electrical and Computer Engineering, Stevens Institute of Technology, Hoboken, US}

\address[h]{Institute of Medical Informatics, University of Luebeck, Luebeck, Germany}


\begin{abstract}
Cervical cancer is the seventh most common cancer among all the cancers worldwide and the fourth most common cancer among women. Cervical cytopathology image classification is an important method to diagnose cervical cancer. Manual screening of cytopathology images is time-consuming and error-prone. The emergence of the automatic computer-aided diagnosis system solves this problem. This paper proposes a framework called CVM-Cervix based on deep learning to perform cervical cell classification tasks. It can analyze pap slides quickly and accurately. CVM-Cervix first proposes a Convolutional Neural Network module and a Visual Transformer module for local and global feature extraction respectively, then a Multilayer Perceptron module is designed to fuse the local and global features for the final classification. Experimental results show the effectiveness and potential of the proposed CVM-Cervix in the field of cervical Pap smear image classification. In addition, according to the practical needs of clinical work, we perform a lightweight post-processing to compress the model.
\end{abstract}

\begin{keyword}
Convolutional Neural Network \sep  Visual Transformer \sep  Multilayer Perceptron	 \sep Cervical cancer \sep Pap smear  \sep Image classification 
\end{keyword}

\end{frontmatter}


\section{Introduction\label{sec:Introduction}}
Cervical cancer is the seventh most common cancer among all the cancers worldwide and the fourth most common cancer among women. In underdeveloped countries, the mortality rate from cervical cancer is very high. Premature sexual intercourse, intercourse with multiple partners, early pregnancy, weak immune system, smoking, having oral contraceptives, and improper menstrual hygiene are the leading causes of cervical cancer. The common symptoms of cervical cancer are abnormal vaginal bleeding, vaginal discharge, and moderate pain during sexual intercourse~\cite{vsarenac2019cervical}. If it can be detected early and given the proper treatment, cervical cancer can be prevented.

Cytopathology screening is widely practised to diagnose cervical malignancies. The procedure of cervical cytopathological examination is that a doctor collects cells on a patient’s cervix using brushes and then places the exfoliated cells on a glass slide. Cytopathologists observe under the microscope to determine whether there is a malignant tumor. However, there are thousands of cells on each slide~\cite{GENCTAV2012Unsupervised}. Therefore, manual inspection is very troublesome, and experts are prone to make mistakes. So this problem needs a better solution.

The emergence of the automatic computer-aided diagnosis system (CAD) solves this problem. CAD can analyze pap slides quickly and accurately. Early CAD was based on classical image processing and traditional machine learning methods, where the manually designed features may lead to unsatisfactory classification results~\cite{goceri2018biomedical}. Nowadays, deep learning technology is relatively mature, and it makes achievements in computer vision, medical imaging, Natural Language Processing (NLP), and other fields. Deep learning technology is different from classical machine learning, and it can automatically learn features for classification~\cite{lecun2015deep}. However, training a deep learning model requires a large number of labelled data sets.

Convolutional Neural Network (CNN) is currently a popular deep learning network, widely used in image segmentation, detection and classification~\cite{goceri2021diagnosis,gocceri2021application,gocceri2020impact}. Recently, Visual Transformer (VT), which originally uses Transformer to process NLP tasks, appears in the field of computer vision. The VT model that first appeared in the field of computer vision is Vision Transformer (ViT). On the premise of pre-training with a large amount of data, it is better than the latest convolutional network, and the computing resources required for training are greatly reduced~\cite{Dosovitskiy2020An}.

In this study, we introduce CVM-Cervix that uses CNN, VT and Multilayer Perceptron (MLP). We have combined SIPaKMeD and CRIC datasets, consisting of 11 categories of single-cell cervical cytopathological images~\cite{plissiti2018sipakmed,rezende2021cric}. CVM-Cervix achieves a classification accuracy of 91.72\%. The workflow of CVM-Cervix is shown in Fig.~\ref{FIG:workflow}. From the workflow diagram, it can be seen that:

\begin{figure}[h]
	\centering
	  \includegraphics[scale=.6]{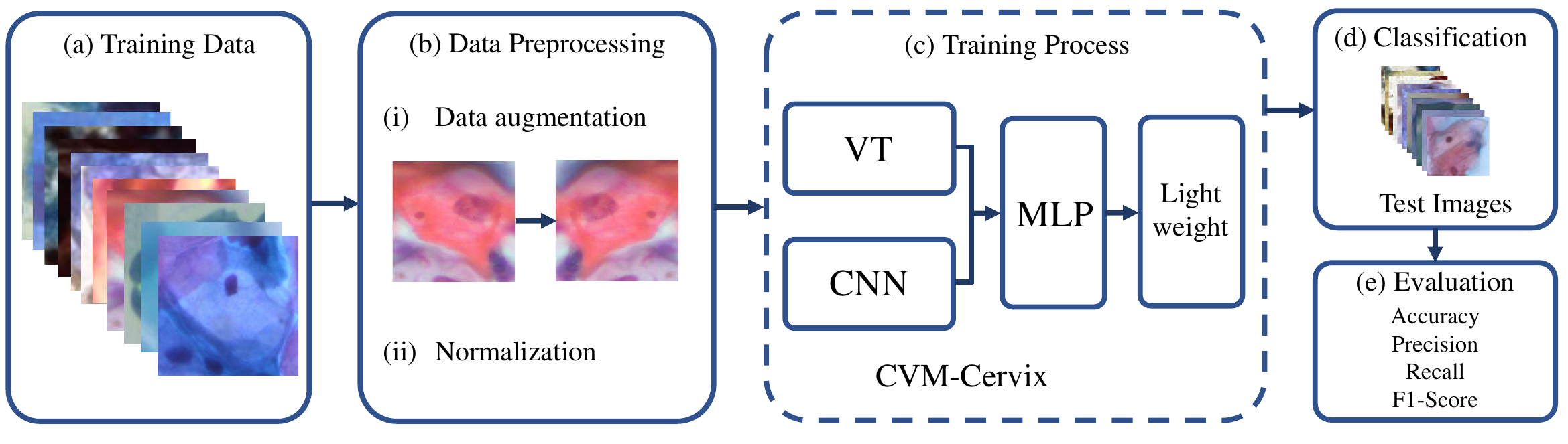}
	\caption{\centering{Workflow of the proposed CVM-Cervix model.}}
	\label{FIG:workflow}
\end{figure}

\begin{itemize}

\item[(a)] 
Cervical cytopathological images are obtained from SIPaKMeD and CRIC datasets and used as the training data. Sec.~\ref{sec:Experiments and Analysis} explains the details of the dataset.

\item[(b)] 
In the preprocessing stage, the training datasets are augmented and normalized.

\item[(c)] 
In this stage, the preprocessed data are fed into CNN and VT modules to extract local and global features. Then, the extracted features are input into MLP module for fusion and classification. Furthermore, we introduce a lightweight post-processing to compress the model.

\item[(d)] 
In this stage, unseen test images are supplied to CVM-Cervix for classification.

\item[(e)] 
Finally, the performance of CVM-Cervix is evaluated by calculating the precision, recall, F1-Score, and accuracy.
\end{itemize}

The main contributions of this paper are as follows: (1) This study performs a 11-class classification task, and it is the work with the most complex data and the most types in the existing literature. (2) A new method (CVM-Cervix) is proposed, which uses CNN, VT and MLP modules for feature extraction and fusion to improve the overall classification performance. In addition, a lightweight post-processing is used to compress the model. (3) The performance of our CVM-Cervix exceeds that of existing methods and achieves a good result of 91.72\%.

This paper structure is as follows: Sec.~\ref{sec:Related Work} elaborates the application and related methods of deep learning in cervical cell image classification tasks. Sec.~\ref{sec:Method} describes our proposed CVM-Cervix. Sec.~\ref{sec:Experiments and Analysis} explains the dataset, experimental results, experimental environment, and evaluation methods. Sec.~\ref{sec:Discussion} discusses some issues in our proposed CVM-Cervix. Sec.~\ref{sec:Conclusion and Future Work} summarizes this paper and proposes future work directions.

\section{Related work\label{sec:Related Work}}

\subsection{Deep neural networks}

This subsection provides a brief introduction to some common deep neural networks.

There are many classic CNN models. Xception network introduces depthwise separable convolutions, which improves efficiency. This method separates spatial convolution and channel convolution~\cite{Chollet2017Xception}. VGG network uses deeper networks and smaller filters. This work uses a 3 $\times$ 3 filter, which has the same acceptance field as a 7 $\times$ 7 filter, but the amount of parameters is greatly reduced~\cite{simonyan2014very}. ResNet network introduces residual blocks. The residual network can turn the deep network into a shallow network by setting other layers in the deep network as identity mapping~\cite{he2016deep}. DenseNet network introduces direct connections between any two layers with the same feature map size. DenseNet can be extended to hundreds of layers and still be optimized~\cite{huang2017densely}. In~\cite{wang2018scene}, an end-to-end attentional recurrent convolutional network is proposed. The method can learn to focus on some important locations and process them on high-level features, discarding unimportant information to improve classification accuracy. In~\cite{wang2020looking}, a multi-scale scene classification method based on a two-stream architecture is proposed. Global and local streams of this method can extract global and local features of the image, respectively. This method proposes a key region detection strategy to connect global and local streams.

Many Visual Transformer models appear. ViT model applies Transformer in the field of NLP to the field of computer vision. Under the training of a large amount of data, the effect of ViT is better than CNN~\cite{Dosovitskiy2020An}. BoTNet combines self-attention and ResNet. While obtaining local information, it also can obtain global information. While reducing the number of parameters, the effect is better than ResNet~\cite{Srinivas2021Bottleneck}. DeiT introduces distillation tokens, which are added after the image block sequence. The training of DeiT requires fewer data and fewer computing resources~\cite{touvron2020training}. T2T-ViT proposes a new T2T mechanism based on ViT, which overcomes the shortcomings of the original simple tokenization. T2T-ViT requires fewer parameters than ViT, but the effect is better~\cite{yuan2021tokens}.

%

\subsection{Applications of deep learning in cervical cell image classification tasks}

This subsection introduces applications of deep learning in cervical cytopathology image classification. Please follow our review paper about cervical cytopathology image analysis using deep learning for further information~\cite{Rahaman2020A}.


In~\cite{wieslander2017deep}, the performances of the ResNet and VGG networks are evaluated. For the cervical dataset, the accuracy rate varies between 84-86\%. The results show that ResNet has higher accuracy and smaller standard deviation, which is a better network. In~\cite{mamunur2021deepcervix}, a hybrid deep feature fusion method based on deep learning is proposed. This method uses pre-trained VGG16, VGG19, Xception, and ResNet50 to extract features and uses a feature fusion technology. This method achieves an accuracy of 99.85\% for a binary classification task of the SIPaKMeD dataset. A method for cervical cell classification based on graph convolutional network (GCN) is proposed~\cite{shi2021cervical}. This method uses the CNN features of the image for clustering and then constructs the graph structure. GCN can generate relationship-aware features, which are merged into CNN features to enhance feature expressiveness. This method has an accuracy of 98.37\% for a 5-class task.

As can be seen from the references, most studies use the CNN model for classification or feature extraction. None of the studies uses CNN and VT models for cervical cell classification tasks. The CNN model can only extract the local features of the image through the convolution kernel, and the VT model can extract the global features of the image through the attention mechanism. We combine local and global features of the image for better analysis. Therefore, this research is noble and significant in this field.

\section{Method\label{sec:Method}}

\subsection{	CVM-Cervix}

The CVM-Cervix framework is designed to classify cervical cell images and Fig.~\ref{FIG:network} is a structural diagram of CVM-Cervix:

\begin{figure}[h]
	\centering
	  \includegraphics[scale=.6]{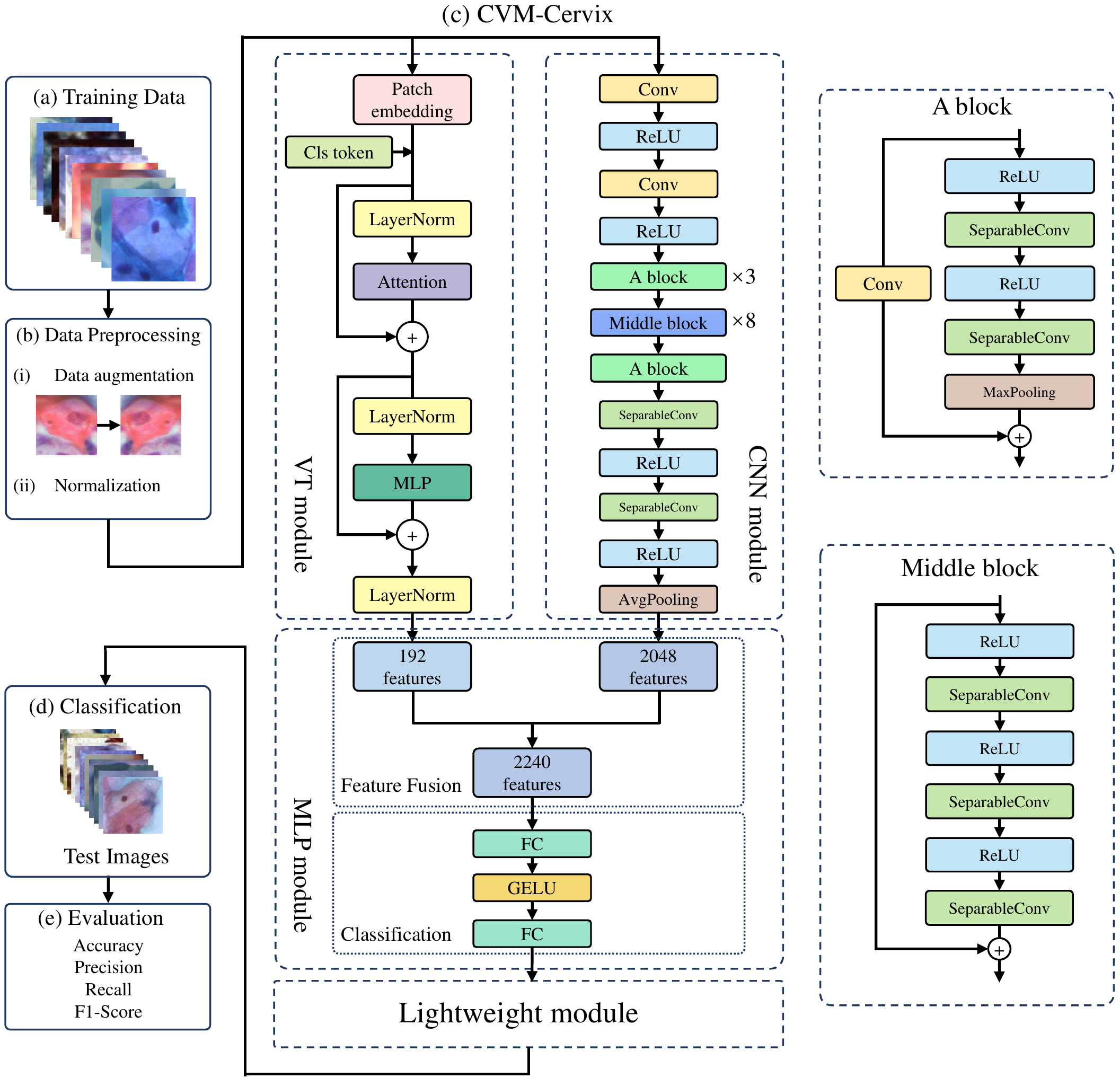}
	\caption{\centering{The structure of CVM-Cervix.}}
	\label{FIG:network}
\end{figure}

\begin{itemize}

\item
Fig.~\ref{FIG:network} (a) shows the cervical cell images used to train CVM-Cervix.

\item
Fig.~\ref{FIG:network} (b) demonstrates the data preprocessing stage. In the preprocessing stage, the first step is data augmentation: first, the input images are resized to 224 $\times$ 224 pixels. Then the random horizontal flip function is used, and the image has a 50\% probability of being flipped horizontally, which can enhance the generalization ability of CVM-Cervix. The second step is to normalize the training data. The training data are input into CVM-Cervix after the preprocessing stage.

\item
Fig.~\ref{FIG:network} (c) shows the main body of CVM-Cervix, which consists of three parts. First, CVM-Cervix designs VT and CNN modules for global and local feature extraction. Then, a MLP module is introduced for feature fusion and classification~\cite{Chollet2017Xception,touvron2020training}. In VT module, we use a tiny DeiT model without distillation tokens, which is equivalent to the ViT model with improved parameters. This module retains all the structure of the DeiT model except the final classification layer, and finally, 192 features are extracted. CNN module uses the official model of Xception, and also retains all the structure except the final classification layer. CNN module finally extracts 2048 features. 
In MLP module, the 192 and 2048 features extracted by VT and CNN modules are first fused to generate 2240 features. Then, MLP is used for classification. MLP consists of two fully connected layers containing the GELU activation function. Furthermore, we post-process the model using a lightweight module to achieve compression.

\item
In Fig.~\ref{FIG:network} (d), we use CVM-Cervix to classify the invisible test images.

\item
In Fig.~\ref{FIG:network} (e), the performance of CVM-Cervix is evaluated by calculating precision, recall, F1-Score, and accuracy.

\end{itemize}

\subsection{CNN module for local feature extraction}

\begin{figure}[h]
	\centering
	  \includegraphics[scale=0.6]{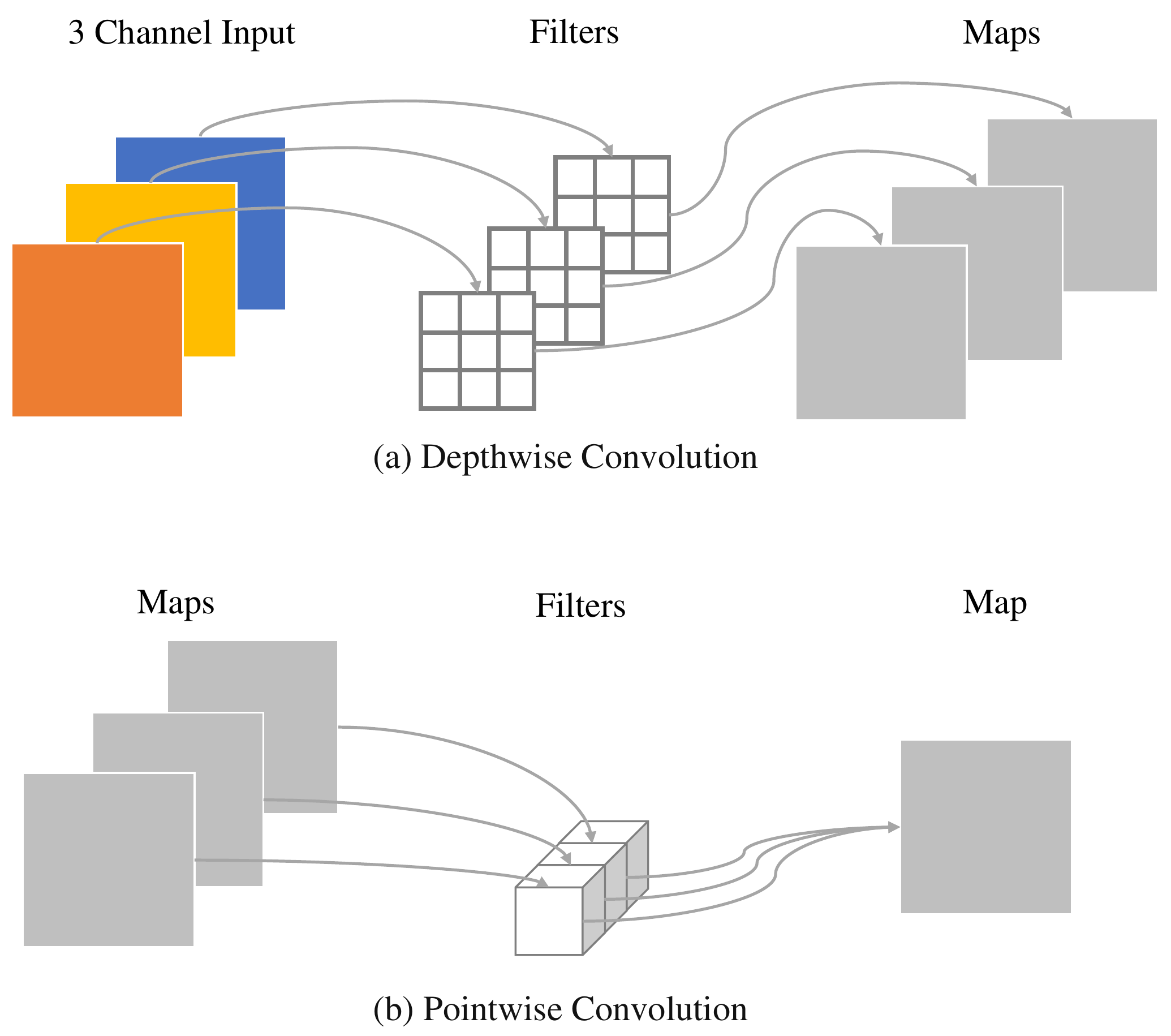}
	\caption{The structure of depthwise separable convolution.}
	\label{FIG:Depthwise_Separable}
\end{figure}

In CVM-Cervix, CNN module is designed to extract local features of cervical cell images. We use the structure other than the fully connected layer in the Xception model as CNN module to extract 2048-dimensional features. The basic components of CNN module are the convolutional layer and the pooling layer. The convolutional layer extracts image features through the function of the convolution kernel. There are two types of pooling layers: the maximum pooling layer and the average pooling layer, which reduce the number of features by looking for the maximum or average value. The advantage of this CNN module is the use of depthwise separable convolution, which is a combination of depthwise convolution and pointwise convolution. Fig.~\ref{FIG:Depthwise_Separable} shows the structure of depthwise separable convolution. Depthwise separable convolution has a lower parameter number and operation cost compared with the conventional convolution operation.

\subsection{VT module for global feature extraction}

In CVM-Cervix, VT module is introduced for global feature extraction in cervical cell images. We use the tiny DeiT model without distillation token as VT module to extract 192-dimensional features. The tiny DeiT model without distillation token is equivalent to the ViT model with improved parameters. VT module converts the image into a patch sequence to process 2D images. Then, the patch sequence is flattened and mapped to the required dimension by linear projection. In addition, positional information and classification headers are added to facilitate learning. This sequence is input into the Transformer encoder. The Transformer encoder contains multi-head self-attention and MLP blocks. There is a LayerNorm before each block, and residual connection is applied after each block.

\subsection{MLP module for feature fusion and classification}	

In CVM-Cervix, MLP module is used for feature fusion and classification, where an early fusion strategy is applied. The traditional methods of early fusion include serial and parallel fusion. We chose the serial fusion method. If the dimensions of the two input features are $m$ and $n$, then the dimension of the output feature is $m+n$. Therefore, the fusion process is to concatenate the 2048-dimensional features extracted by CNN module and the 192-dimensional features extracted by VT module to obtain 2240-dimensional features.

Furthermore, a MLP classifier is designed for the final classification. MLP is the most brief and effective artificial neural network algorithm, and it is a feedforward neural network. It learns through backpropagation technology. MLP has an input layer, a hidden layer, and an output layer. The input layer is the receiver of MLP. There can be one or more hidden layers in the middle, which are used to calculate data and iterate. The output layer can predict the classification result. The simplest MLP contains only one hidden layer, which is a three-layer structure. The use of the activation function can introduce nonlinearity to the neural network and enhance the expressive ability of the network. The MLP here is two fully connected layers containing GELU activation functions, which are used to classify 2240-dimensional features.

\section{Experiments and analysis\label{sec:Experiments and Analysis}}

\subsection{Datasets}

\subsubsection{CRIC}

\begin{figure}[h]
	\centering
	  \includegraphics[scale=0.54]{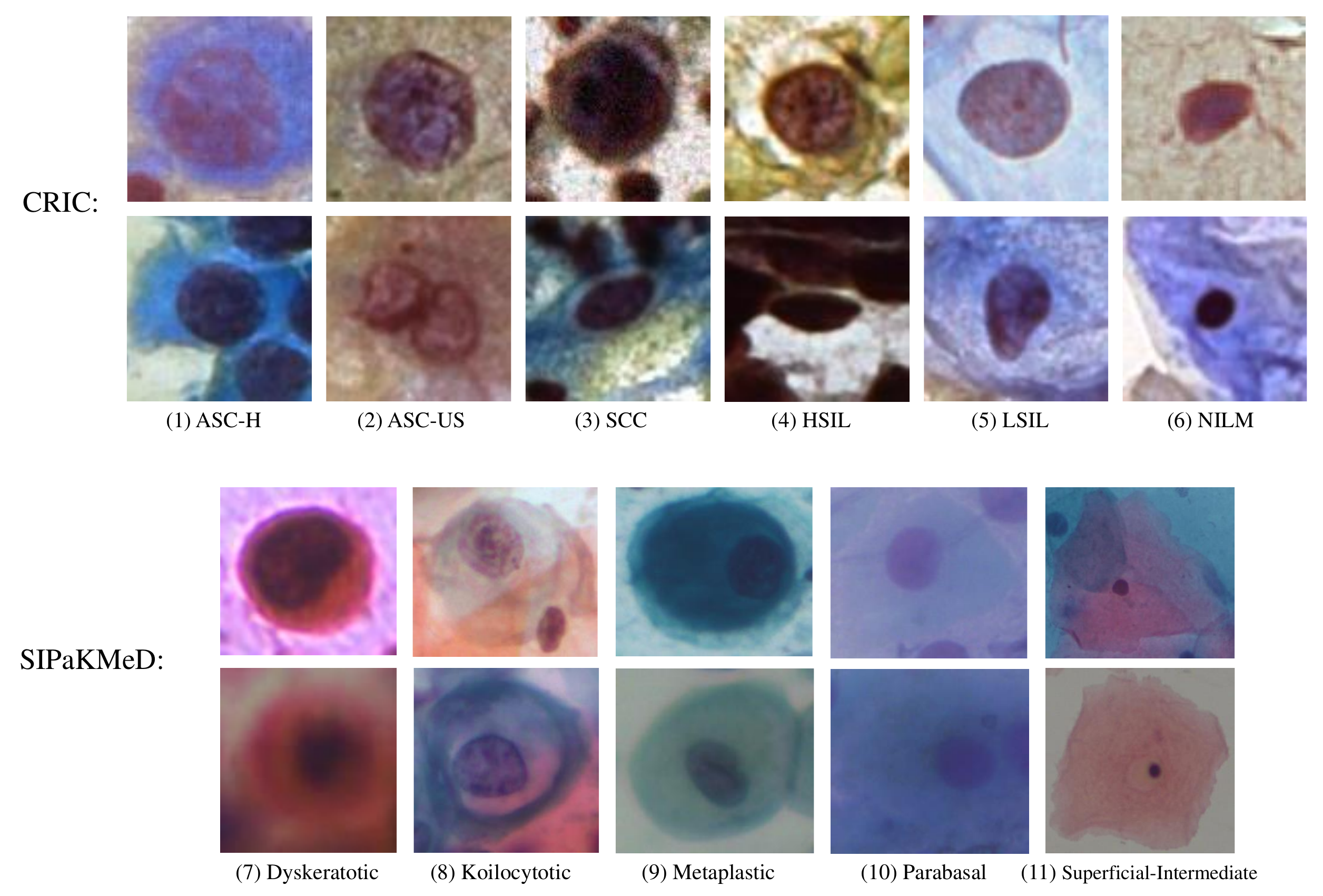}
	\caption{An example of images in CRIC and SIPaKMeD datasets.}
	\label{FIG:SIPaKMeD_and_CRIC}
\end{figure}

The CRIC dataset has 400 images from traditional Pap smears, including 11534 cells manually classified~\cite{rezende2021cric}. It is classified by several cytopathologists according to the Bethesda system nomenclature. There are six types of images: atypical squamous cells, cannot exclude a high-grade lesion (ASC-H); atypical squamous cells of undetermined significance, possibly non-neoplastic (ASC-US); squamous cell carcinoma (SCC); high-grade squamous intraepithelial lesion (HSIL); low-grade squamous intraepithelial lesion (LSIL); negative for intraepithelial lesion or malignancy (NILM)~\cite{rezende2021cric}. In this paper, only cropped images are used, including 4789 images of separated cervical cells~\cite{n2021deep}. Some examples can be seen in Fig.~\ref{FIG:SIPaKMeD_and_CRIC}.

\subsubsection{SIPaKMeD}

The SIPaKMeD dataset can be used for cervical cell image classification tasks. This dataset contains 4049 cervical cell images cropped from the clustered cell images, which are obtained by the CCD camera. There are five categories of these images: dyskeratotic, koilocytotic, metaplastic, parabasal, and superficial intermediate~\cite{plissiti2018sipakmed}. Fig.~\ref{FIG:SIPaKMeD_and_CRIC} shows some examples.

\subsubsection{Combined dataset}

In this study, we have combined the CRIC and SIPaKMeD datasets and used them in this experiment. Therefore, we obtained 11 categories, including six categories of the CRIC dataset and five categories of the SIPaKMeD dataset. This combined dataset has a total of 8838 images. We randomly selected 60\% of the images in each class to serve as the training set, 20\% as the validation set, and 20\% as the test set. The data settings are shown in Table~\ref{data_arrangement}.

\begin{table}[h]
\caption{Data settings of the combined dataset.}\label{data_arrangement}
\centering

\begin{tabular}{@{} lcccccc@{} }
\toprule
Class/Dataset & Train & Validation & Test & Total \\
\midrule
(1) ASC-H         & 484   & 161        & 161  & 806   \\
(2) ASC-US        & 446   & 148        & 148  & 742   \\
(3) SCC         & 413   & 137        & 137  & 687   \\
(4) HSIL         & 525   & 175        & 174  & 874   \\
(5) LSIL         & 491   & 164        & 163  & 818   \\
(6) NILM         & 518   & 172        & 172  & 862   \\
(7) Dyskeratotic         & 488   & 163        & 162  & 813   \\
(8) Koilocytotic         & 495   & 165        & 165  & 825   \\
(9) Metaplastic         & 476   & 159        & 158  & 793   \\
(10) Parabasal         & 473   & 157        & 157  & 787   \\
(11) \small Superficial-Intermediate        & 499   & 166        & 166  & 831   \\

Total      & 5308  & 1767       & 1763 & 8838 \\
\bottomrule
\end{tabular}
\end{table}

\subsection{Experimental environment}

The experiments in this paper are carried out on a local workstation with 32GB RAM and the windows 10 operating system. The GPU of the workstation is an 8GB NVIDIA Quadro RTX 4000. We use the Python programming language, and the version of Pytorch is 1.8.0. When training CVM-Cervix, the learning rate is set to 0.0002, the batch size of the training set is 16, the epoch is 100, and the AdamW optimizer is used.

\subsection{Evaluation methods}

Precision, recall, F1-Score, and accuracy are calculated to evaluate the performance of CVM-Cervix. True Positive (TP) is the number of positive samples predicted accurately. True Negative (TN) is the number of negative samples predicted accurately. The number of negative samples predicted to be positive is False Positive (FP). And the number of positive samples predicted to be negative is False Negative (FN). The proportion of TP in all positive predictions is precision. A recall is the ratio of predicted positive samples to the total positive sample numbers. F1-Score is a comprehensive indicator of precision and recall. Accuracy is the ratio of the number of correct predictions to the total. The formulas are shown in Table~\ref{tbl3}.

\begin{table}[h]
\caption{Evaluation metrics}\label{tbl3}
\renewcommand\arraystretch{1.5}
\centering

\begin{tabular}{@{} cccc@{} }
\toprule
Assessment      & Formula                  & Assessment & Formula                               \\
\midrule
Precision ($P$) & $\rm \frac{TP}{TP + FP}$ & F1-Score   & $ 2 \times \frac{P \times R}{P + R}$  \\
Recall ($R$)    & $\rm \frac{TP}{TP + FN}$ & Accuracy   & $\rm \frac{TP+TN}{TP + TN + FP + FN}$\\
\bottomrule
\end{tabular}
\end{table}

\subsection{Experimental results and analysis}

\begin{figure}
	\centering
	  \includegraphics[scale=0.6]{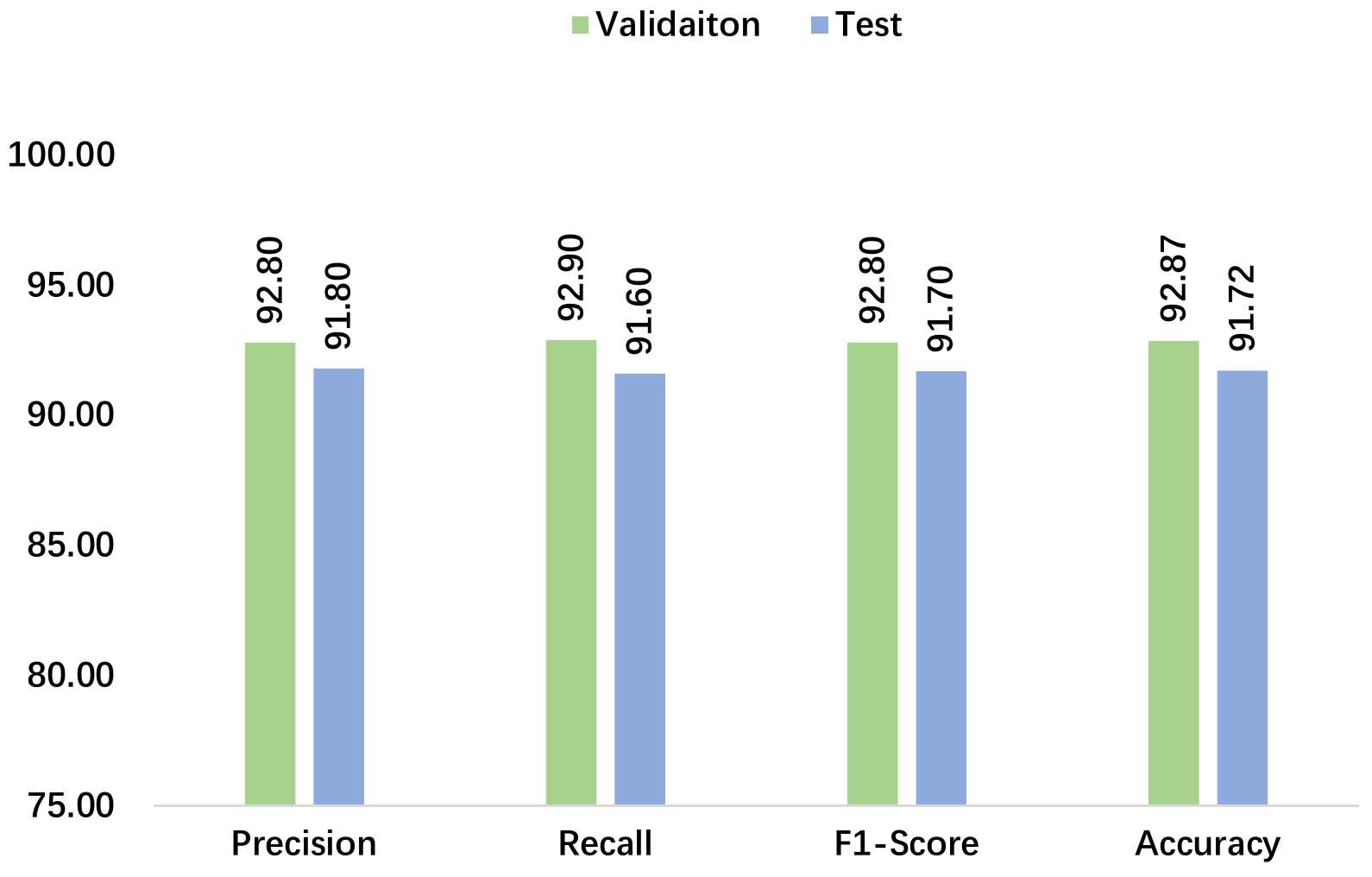}
	\caption{The performance of CVM-Cervix.}
	\label{FIG:model_performence}
\end{figure}

In the validation and test phases, the precision, recall, F1-Score, and accuracy are calculated to evaluate the performance of the proposed CVM-Cervix model, as shown in Fig.~\ref{FIG:model_performence}. It can be seen that the precision, recall, F1-Score, and accuracy of CVM-Cervix on the validation set are 92.80\%, 92.90\%, 92.80\%, and 92.87\%, respectively. The precision, recall, F1-Score, and accuracy of CVM-Cervix on the test set are 91.80\%, 91.60\%, 91.70\%, and 91.72\%, respectively. The performance difference of CVM-Cervix on the validation set and the test set is about 1\%, which proves its strong generalization ability. The performance of CVM-Cervix surpasses the existing methods and achieves very good results. 

\begin{figure}
	\centering
	  \includegraphics[scale=0.46]{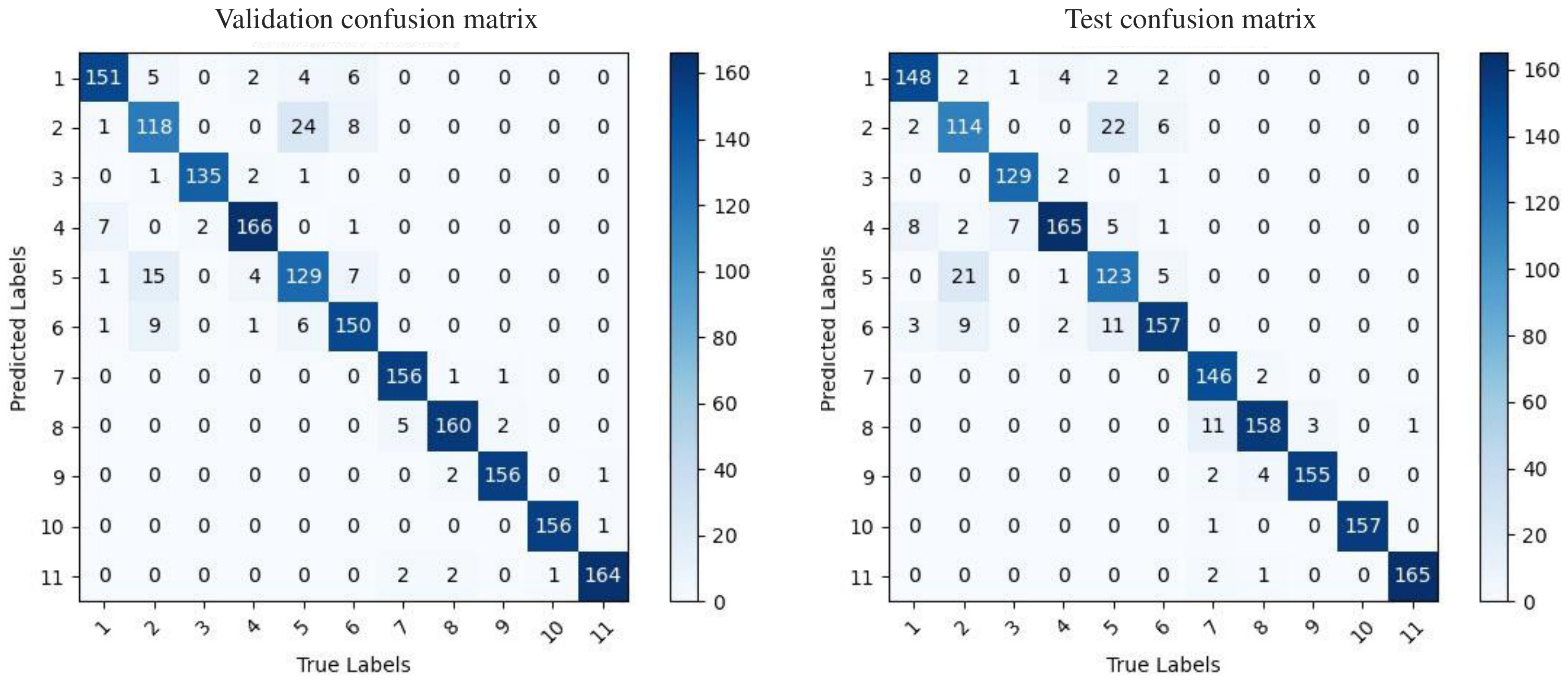}
	\caption{The confusion matrices of the validation and test results using CVM-Cervix.}
	\label{FIG:conf_matrix}
\end{figure}

The confusion matrices of the validation and test results using CVM-Cervix are shown in Fig.~\ref{FIG:conf_matrix}. In the validation confusion matrix, we can see that the classifications of the first, the third, and the sixth to the eleventh categories are very accurate. Fifteen images of the second category are classified into the fifth category, and nine images are classified into the sixth category. Twenty-four images of the fifth category are classified into the second category. The situation is roughly the same on the test confusion matrix. Twenty-one images of the second category are misclassified into the fifth category, and nine images are misclassified into the sixth category. Twenty-two images of the fifth category are misclassified into the second category, and 11 images are misclassified into the sixth category. In the seventh category, 11 images are misclassified into the eighth category. 

Among these eleven categories, the first, third, fourth, seventh, and eighth types are abnormal types. It can be observed that CVM-Cervix has a higher classification accuracy for abnormal types. We hope that the model will not misclassify cancer patients in practical applications, because this is a matter of life and death. Therefore, the higher the accuracy of the model's classification of abnormal types, the better, which is exactly the performance of our CVM-Cervix.

\subsection{Computational time}

In this experiment, we use 100 epochs to train CVM-Cervix, and the model parameter size is about 120 MB. There are 5308 training images, and the total training time is 2.3 hours. The training time for each epoch is about 82 seconds. Although the training time is long, the test time is only 31 seconds. There are 1763 test images, and the test time for each image is about 0.018 seconds.

\subsection{Lightweight post-processing}

In addition, according to the practical needs of clinical work, we perform a lightweight post-processing to compress the model~\cite{choudhary2020comprehensive}. Firstly, we use a quantization technique to reduce the number of bits required to store each weight from 32 to 16 bits. Then, the model parameter size is reduced to 60 MB. Finally, the classification performance is almost unchanged (around -1\%), and the total training time is shortened (around 1 hour).

\subsection{Extended experiment}

\subsubsection{Extended experiment on the Combined dataset: comparative experiment\label{sec:Comparative Experiment}}

To see the performance of CVM-Cervix, 22 models are selected for this comparative experiments, including 18 CNN models: VGG11~\cite{simonyan2014very}, VGG13~\cite{simonyan2014very}, VGG16~\cite{simonyan2014very}, VGG19~\cite{simonyan2014very}, ResNet18~\cite{he2016deep}, ResNet34~\cite{he2016deep}, ResNet50~\cite{he2016deep}, ResNet101~\cite{he2016deep}, DenseNet121~\cite{huang2017densely}, Dense\-Net169~\cite{huang2017densely}, InceptionV3~\cite{Szegedy2016Rethinking}, Xception~\cite{Chollet2017Xception}, AlexNet~\cite{russakovsky2015imagenet}, GoogLeNet~\cite{Szegedy2015googlenet}, MobileNetV2~\cite{sandler2018mobilenetv2}, ShuffleNetV2$\times$1.0~\cite{ma2018shufflenet}, ShuffleNetV2$\times$0.5~\cite{ma2018shufflenet}, InceptionResNetV1~\cite{szegedy2017inception}. The remaining four are VT models: ViT~\cite{Dosovitskiy2020An}, BoTNet~\cite{Srinivas2021Bottleneck}, DeiT~\cite{touvron2020training}, T2T-ViT~\cite{yuan2021tokens}.

The experimental settings are the same. Table~\ref{compare_expreriment} shows the results of models on the test set of the combined dataset. It can be seen that DenseNet169 obtains the highest accuracy rate of 88.99\% in CNN models. In VT models, DeiT has the highest accuracy rate of 87.35\%. Among all the models, the proposed CVM-Cervix achieves the highest test accuracy of 91.72\%, which is nearly 3\% higher than the accuracy of DenseNet169 and DeiT. In addition, the precision, recall, and F1-Score of CVM-Cervix are also the highest, which are 91.80\%, 91.60\%, and 91.70\%, respectively. This shows the effectiveness of the proposed CVM-Cervix model.

\begin{table}
\caption{Comparison results of models on the test set of the combined dataset.}\label{compare_expreriment}
\centering
\scalebox{1.0}{
\begin{tabular}{@{} ccccc@{} }
\toprule
Models            & Avg. P & Avg. R & Avg. F1 & Acc (\%)  \\
\midrule
VGG11             & 84.80  & 83.90  & 84.00   & 84.06    \\
VGG13             & 85.30  & 84.70  & 84.40   & 84.62    \\
VGG16             & 79.80  & 79.90  & 79.60   & 79.97    \\
VGG19             & 78.60  & 78.30  & 78.20   & 78.38    \\
ResNet18          & 88.40  & 88.30  & 88.20   & 88.25    \\
ResNet34          & 88.00  & 88.10  & 87.90   & 88.03    \\
ResNet50          & 87.60  & 87.60  & 87.30   & 87.57    \\
ResNet101         & 86.50  & 86.40  & 86.30   & 86.38    \\
DenseNet121       & 87.90  & 87.80  & 87.70   & 87.86    \\
DenseNet169       & 89.20  & 88.90  & 88.70   & \textbf{88.99}    \\
InceptionV3       & 85.40  & 84.90  & 84.80   & 84.96    \\
Xception          & 87.70  & 87.70  & 87.60   & 87.69    \\
AlexNet           & 86.20  & 86.20  & 86.10   & 86.10    \\
GoogLeNet         & 87.60  & 87.20  & 87.10   & 87.29    \\
MobileNetV2       & 84.90  & 84.60  & 84.60   & 84.62    \\
ShuffleNetV2$\times$1.0  & 85.50  & 84.30  & 84.50   & 84.57    \\
ShuffleNetV2$\times$0.5  & 82.60  & 82.60  & 82.50   & 82.64    \\
InceptionResNetV1 & 85.20  & 84.90  & 84.90   & 84.96    \\
ViT               & 86.10  & 86.00  & 85.70   & 85.87    \\
BoTNet            & 66.10  & 65.50  & 63.90   & 65.51    \\
DeiT              & 87.40  & 87.40  & 87.20   & \textbf{87.35}    \\
T2T-ViT           & 85.60  & 85.30  & 85.20   & 85.19    \\
\textbf{CVM-Cervix}        & \textbf{91.80}  & \textbf{91.60}  & \textbf{91.70}   & \textbf{91.72} \\

\bottomrule
 \end{tabular}}
\end{table}

\subsubsection{Extended experiment on Combined dataset: classification with data augmentation\label{sec:SIPaKMeD_aug_extend}}

\begin{figure}[h]
	\centering
	  \includegraphics[scale=0.6]{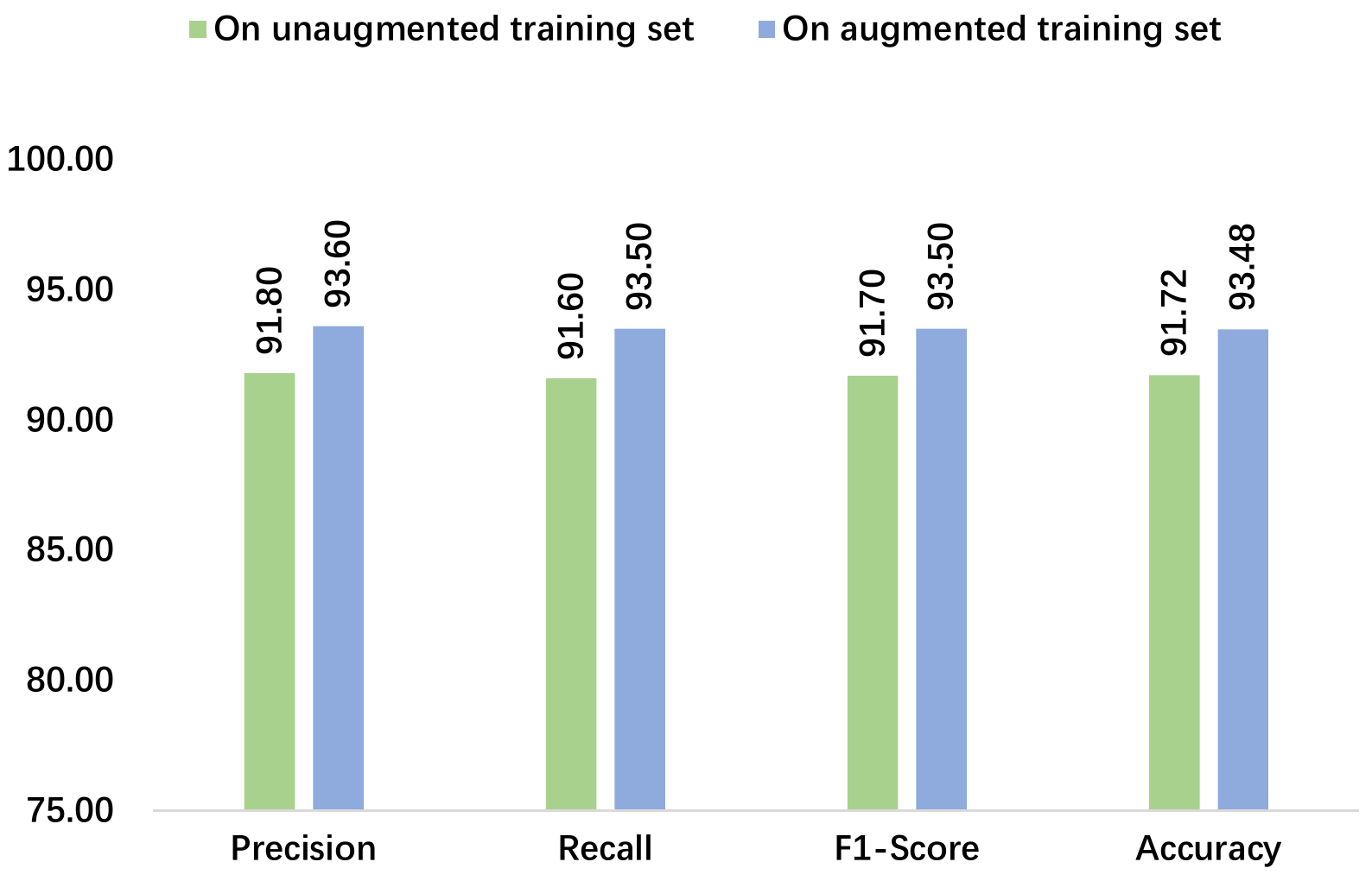}
	\caption{The comparison of test results of CVM-Cervix on the unaugmented and augmented training sets.}
	\label{FIG:augmented_exp}
\end{figure}

In this extended experiment, the training set is augmented. The data increase four times by rotating 180 degrees, flipping left and right, and flipping up and down. Then CVM-Cervix is used on the augmented training set for the experiment. The comparison of test results of CVM-Cervix on the unaugmented and augmented training sets are shown in Fig.~\ref{FIG:augmented_exp}.

It can be seen that the precision, recall, F1-Score, and accuracy of CVM-Cervix on the augmented training set are 93.60\%, 93.50\%, 93.50\%, and 93.48\%, respectively. There is a roughly 2\% improvement compared to the effect on the unaugmented training set. The reason for the improvement is the increase in data so that CVM-Cervix can be better trained. The improvement is not large because the performance of the proposed CVM-Cervix model on the unaugmented training set has stabilized. It also illustrates the superiority of our CVM-Cervix.

\subsubsection{Extended experiment on Herlev dataset \label{sec:Herlev_dataset}}

\begin{figure}[h]
	\centering
	  \includegraphics[scale=0.53]{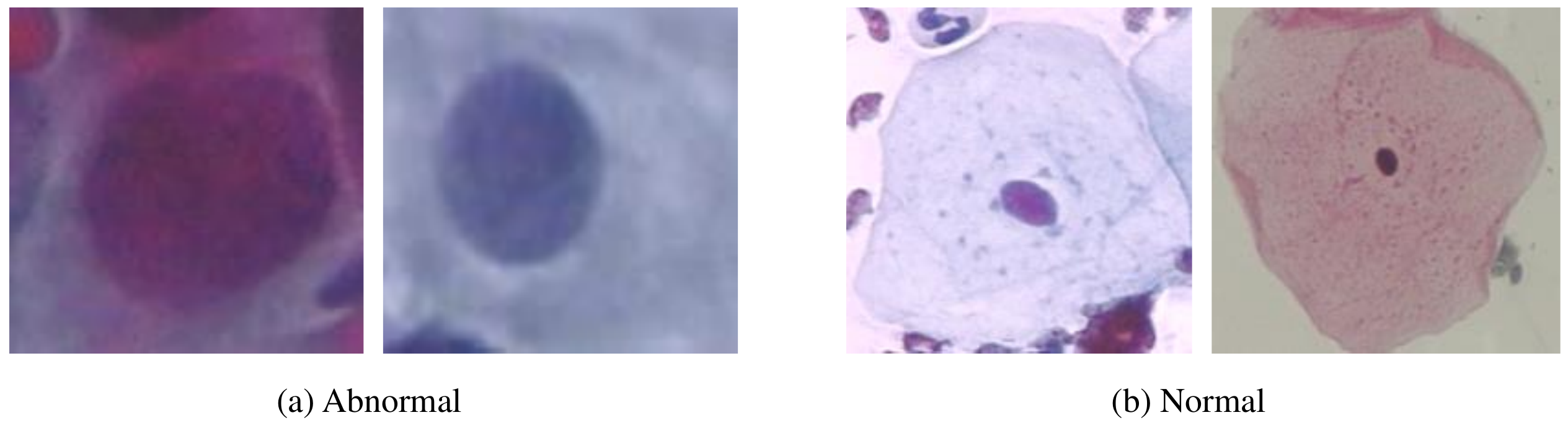}
	\caption{An example of images in Herlev dataset.}
	\label{FIG:Herlev}
\end{figure}

To validate the generalization ability of CVM-Cervix on other cervical Pap smear datasets, we use Herlev dataset for extended experiments~\cite{jantzen2005pap}. The experimental setup is the same as the main experiment. There are two types in the Herlev dataset: Normal and Abnormal. Some examples are shown in Fig.~\ref{FIG:Herlev}. This dataset has 917 images, and we randomly divide the training set, validation set and test set in a ratio of 6:2:2. Due to the small amount of data, we augment the training set 4 times. The data settings are shown in Table~\ref{herlev_data}.

\begin{table}[h]
\caption{Data settings of the Herlev dataset.}\label{herlev_data}
\centering

\begin{tabular}{@{} lcccccc@{} }
\toprule
Class/Dataset & Train & Validation & Test & Total \\
\midrule
Abnormal  & 1620   & 135        & 135  & 1890   \\
Normal     & 584     & 48         & 48     & 680  \\
Total              & 2204   & 183       & 183   & 2570 \\
\bottomrule
\end{tabular}
\end{table}

\begin{figure}
	\centering
	  \includegraphics[scale=0.6]{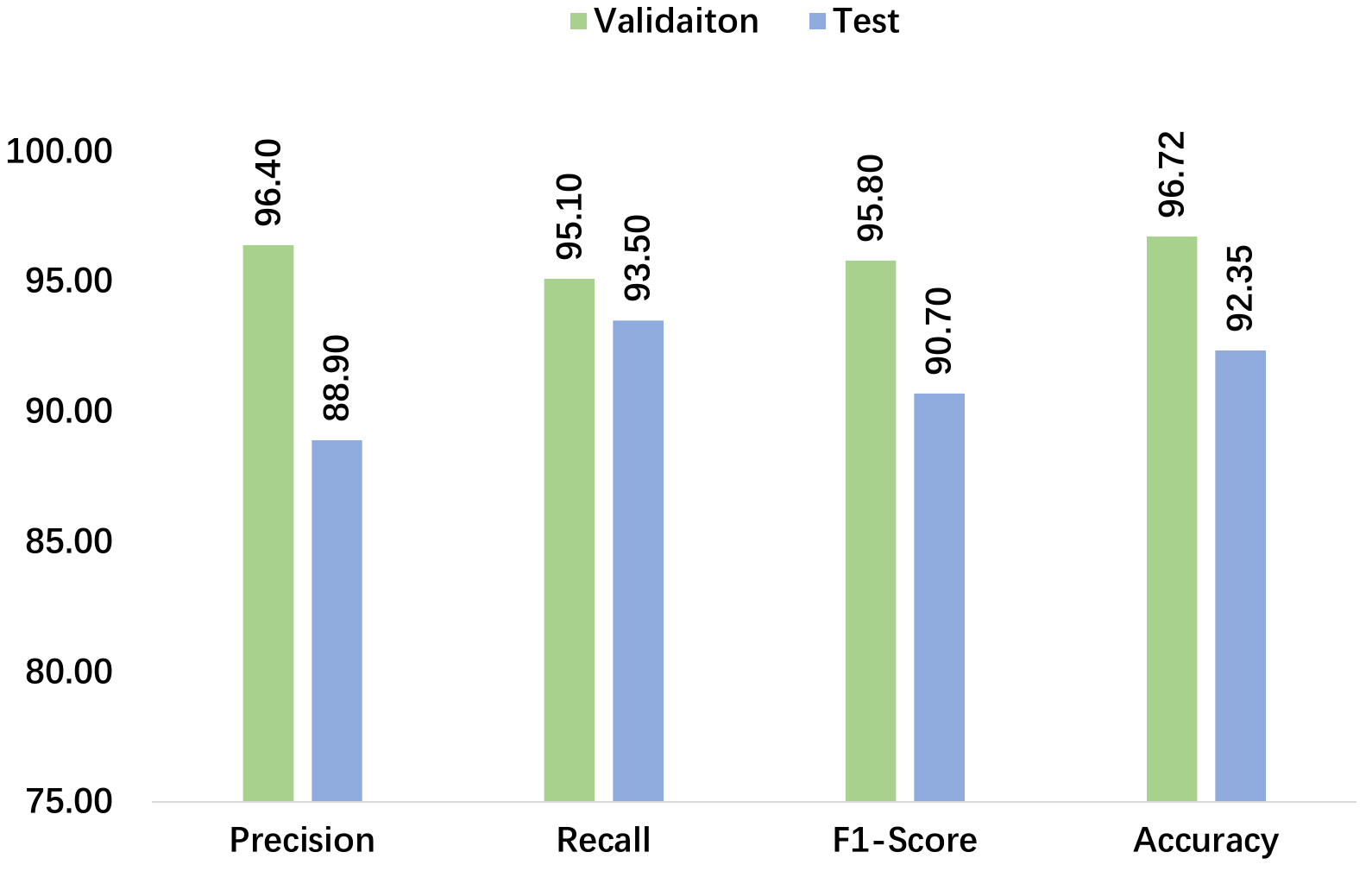}
	\caption{The performance of CVM-Cervix on Herlev dataset.}
	\label{FIG:herlev_result}
\end{figure}

Fig.~\ref{FIG:herlev_result} shows the performance of CVM-Cervix on the Herlev dataset. It can be seen that the precision, recall, F1-Score and accuracy on the validation set are 96.40\%, 95.10\%, 95.80\% and 96.72\%, respectively. The precision, recall, F1-Score and accuracy on the test set are 88.90\%, 93.50\%, 90.70\% and 92.35\%, respectively. CVM-Cervix performs slightly less well on the test set due to too little data and an uneven number of classes, but still good. It indicates that CVM-Cervix generalizes well on other cervical Pap smear datasets.

\subsubsection{Extended experiment on peripheral blood cell dataset \label{sec:peripheral_blood_cell}}

\begin{figure}[h]
	\centering
	  \includegraphics[scale=0.6]{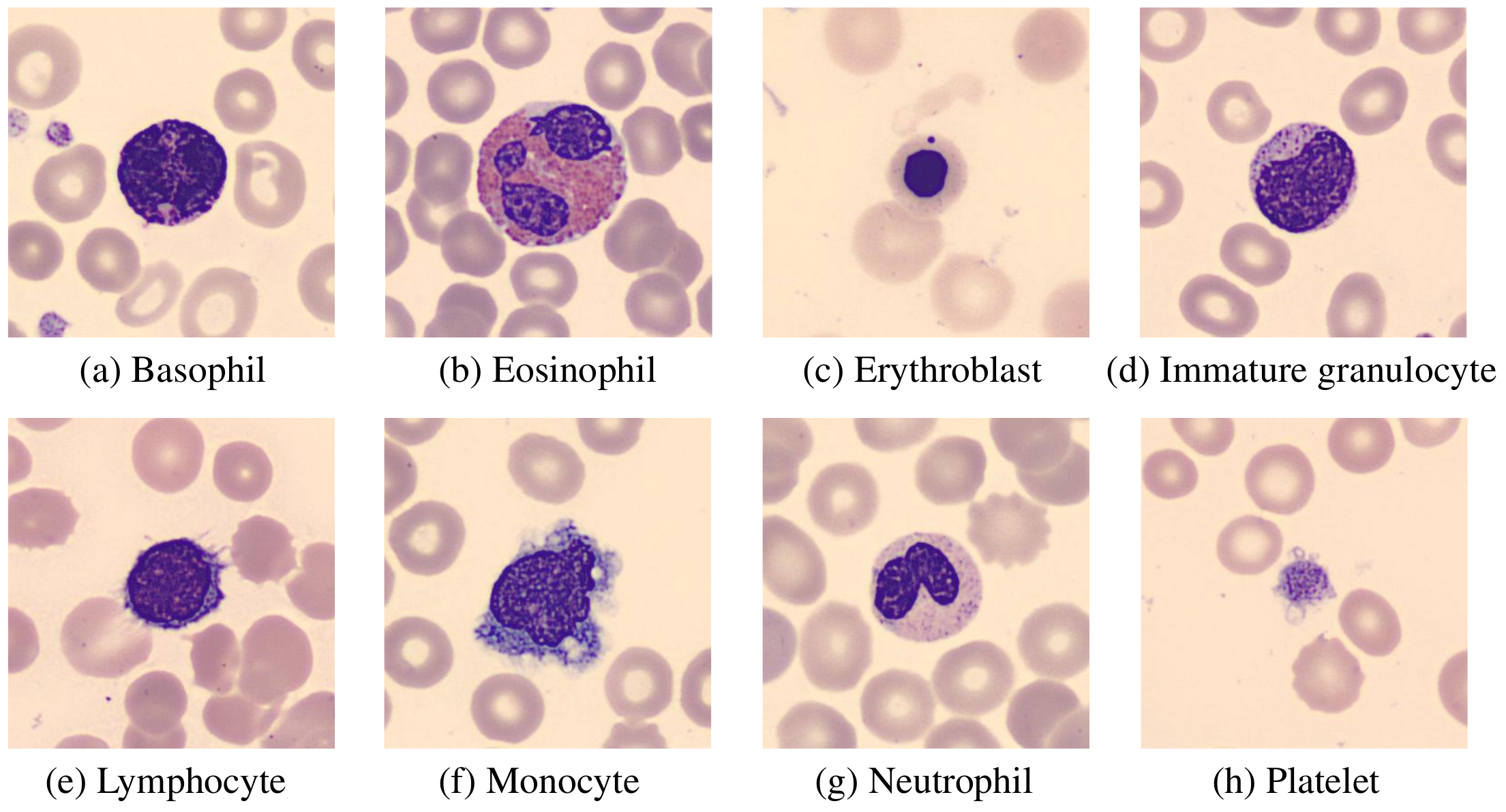}
	\caption{An example of images in peripheral blood cell dataset.}
	\label{FIG:Blood_cell}
\end{figure}

To explore the performance of CVM-Cervix on other cell image datasets, the peripheral blood cell dataset is used for extended experiments~\cite{acevedo2020dataset}. There are 17092 blood cell images in this dataset, with eight types: neutrophil, eosinophil, basophil, lymphocyte, monocyte, immature granulocyte, erythroblast and platelet. Some examples are shown in Fig~\ref{FIG:Blood_cell}. We divide the dataset in the same proportions. Table~\ref{Blood_cell_table} shows the data settings.

\begin{table}[h]
\caption{Data settings of peripheral blood cell dataset.}\label{Blood_cell_table}
\centering
\scalebox{1.0}{
\begin{tabular}{@{} lcccccc@{} }
\toprule
Class/Dataset & Train & Validation & Test & Total \\
\midrule
Basophil        & 731   & 244        & 243 & 1218   \\
Eosinpohil     & 1871     & 623         & 623     & 3117  \\

Erythroblast  & 931   & 310        & 310  & 1551   \\
Immature granulocyte     & 1737     & 579         & 579     &2895  \\

Lymphocyte  & 729   & 243        & 242  & 1214   \\
Monocyte     & 852     & 284         & 284     & 1420  \\

Neutrophil  & 1998   & 666       & 665  & 3329   \\
Platelet     & 1409     & 470         & 469    & 2348  \\

Total         & 10258   & 3419       & 3415   & 17092 \\
\bottomrule
\end{tabular}}
\end{table}

\begin{figure}
	\centering
	  \includegraphics[scale=0.6]{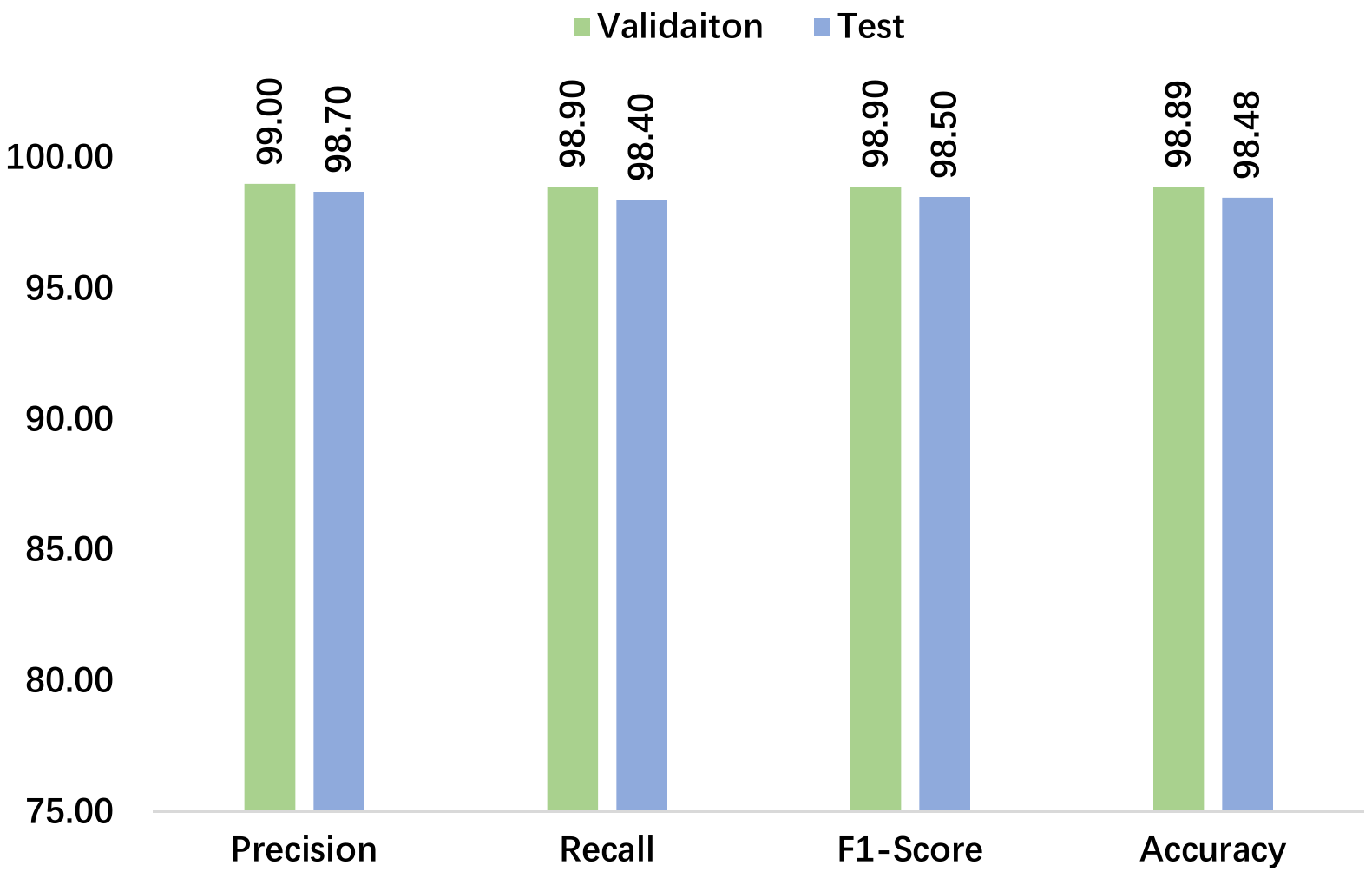}
	\caption{The performance of CVM-Cervix on peripheral blood cell dataset.}
	\label{FIG:bloodcell_result}
\end{figure}

The performance of CVM-Cervix on the peripheral blood cell dataset is shown in Fig.~\ref{FIG:bloodcell_result}. It can be seen that the precision, recall, F1-Score and accuracy on the validation set are 99.00\%, 98.90\%, 98.90\% and 98.89\%, respectively. The precision, recall, F1-Score and accuracy on the test set are 98.70\%, 98.40\%, 98.50\% and 98.48\%, respectively. CVM-Cervix performs very well on both the validation set and the test set of peripheral blood cell images and their difference is small. It indicates that CVM-Cervix performs well on other cell image datasets.

\section{Discussion\label{sec:Discussion}}

\subsection{Misclassification analysis and limitations of CVM-Cervix}

\begin{figure}
	\centering
	  \includegraphics[scale=0.48]{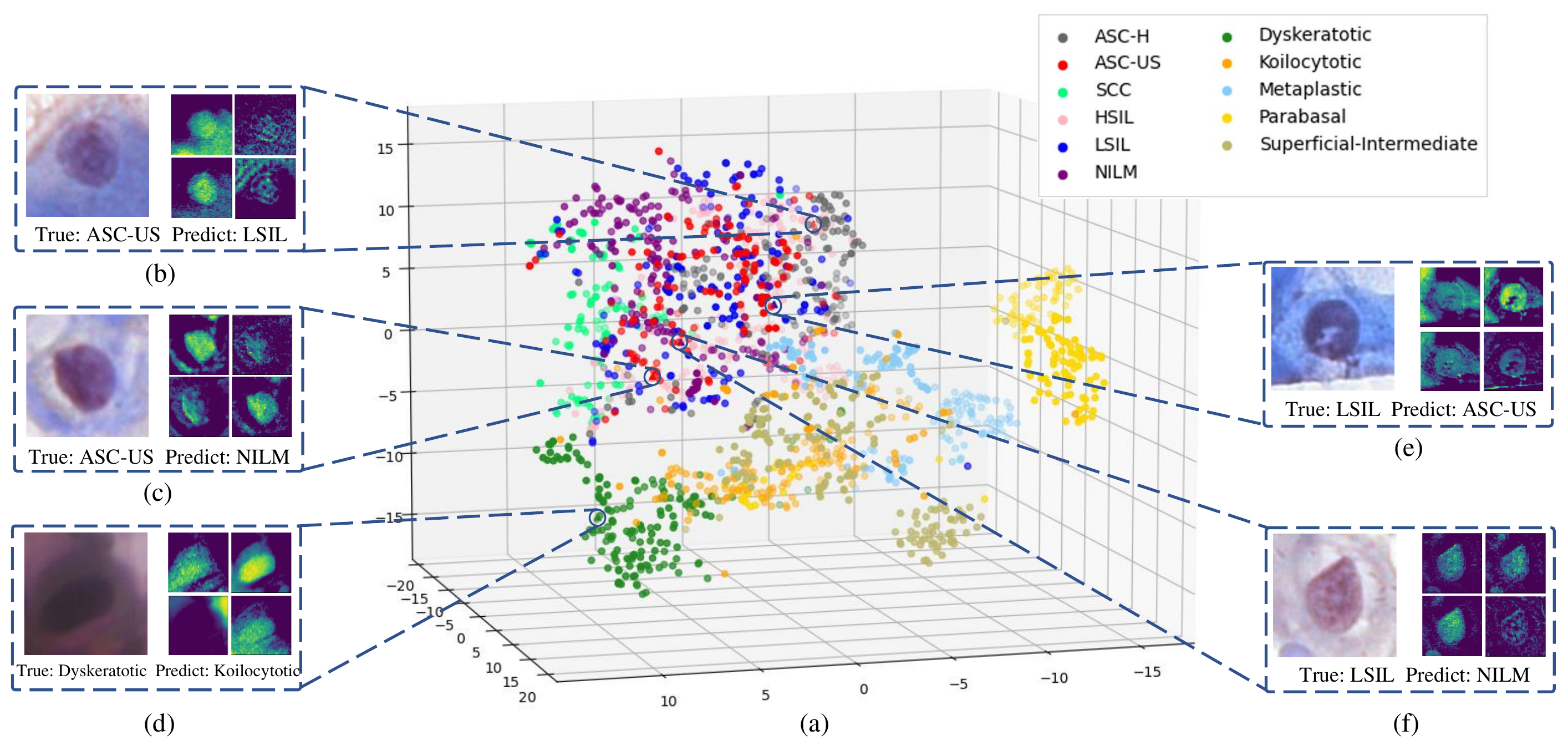}
	\caption{Visual analysis of misclassified images from the combined dataset. (a) is a 3D scatter plot obtained using t-SNE. (b)-(f) are misclassified images and their feature maps.}
	\label{FIG:t_SNE}
\end{figure}

To analyze the reasons for the misclassification, we use t-SNE method to map the 2240-dimensional feature vector into 3-dementioanality (3D) and visualize it~\cite{van2008visualizing}. Fig.~\ref{FIG:t_SNE} (a) shows the 3D scatter plot obtained using t-SNE. Some misclassified images and their feature maps are shown in Fig.~\ref{FIG:t_SNE} (b)-(f).

In Fig.~\ref{FIG:t_SNE} (b) and (c), ASC-US images are classified into LSIL and NILM. In Fig.~\ref{FIG:t_SNE} (e), the LSIL image is classified as ASC-US. In Fig.~\ref{FIG:t_SNE} (f), the LSIL image is classified as NILM. In medicine, these categories are adjacent. It can be seen that the nuclei of the four images are very similar and difficult to distinguish, which is also the main reason for the prediction error. From feature maps, we can see that features extracted by CVM-Cervix are inaccurate due to the light color of the cells and the unclear boundaries. In Fig.~\ref{FIG:t_SNE} (d), the Dyskeratotic image is classified as Koilocytotic. It can be seen that the background of the image is very dark, and the boundary between the nucleus and the cytoplasm is very blurred. The feature map also shows that CVM-Cervix cannot effectively identify cell nuclei when extracting the image features. In general, the similarity and ambiguity of images are the main reasons for classification errors.

Additionally, a limitation of our proposed CVM-Cervix method is that it is currently only performed on the classification task of cell images, and has not been experimented on other tasks.

\subsection{Optimization strategy}

Firstly, in the VT module of CVM-Cervix, we use a tiny DeiT model without distillation tokens. In the original paper of DeiT, the distillation module uses a teacher model to train a student model (ViT)~\cite{touvron2020training}. In contrast, our CVM-Cervix integrates the DeiT module into the framework to use its global feature extraction capabilities and does not require a separate DeiT model. During pre-experiments, we use DeiT with the	 distillation module for experiments. However, we found that DeiT without the distillation module during fusion is more effective for our classification tasks.

Secondly, the optimizer and activation function should be chosen carefully. Although, there are various optimizers (e.g., Sobolev gradient based~\cite{goceri2020capsnet} and Lagrangian optimizer) and activation functions~\cite{goceri2019skin,goceri2019analysis} implemented with deep networks to solve different classification problems, we apply the AdamW and ReLU due to their efficiency in the proposed architecture with our datasets.

Finally, loss functions have an important role in a deep network architecture. To 
improve the performance of CNN-based architectures, hybrid loss functions have been proposed in several works~\cite{goceri2021deep}. We used the default loss function (i.e., cross-entropy) because of its low computational cost and efficiency with our images.

\section{Conclusion and future work \label{sec:Conclusion and Future Work}}

This paper proposes a deep learning-based framework called CVM-Cervix for cervical cell classification tasks. CVM-Cervix first proposes a CNN module and a VT module for local and global feature extraction respectively, then a Multilayer Perceptron module is designed to fuse the local and global features for the final classification. CVM-Cervix is tested on a combined dataset of CRIC and SIPaKMeD datasets, and an accuracy of 91.72\% is obtained for the 11-class classification problem. In addition, according to the practical needs of clinical work, we perform a lightweight post-processing to compress the model, where the model parameter size is compressed from 120 MB to 60 MB. In the extended experiment, CVM-Cervix is more efficient than the CNN and VT models alone. CVM-Cervix performs well on other cervical Pap smear and cell image datasets, indicating its strong generalization ability.

In the future, we will try to use more model combinations to improve CVM-Cervix performance. At the same time, the modification of the module structure can improve the feature extraction capability. Various data preprocessing techniques can be applied in the future to enhance the generalization capabilities of the model, such as color jittering and random cropping.

As another future work, the performance of the proposed approach will be compared with the performance of a capsule network since they can preserve spatial relationships of learned features and have been proposed recently for image classification~\cite{goceri2020capsnet,goceri2021analysis,goceri2021capsule}.

\section*{Acknowledgements and Conflict of Interest}
This work is supported by National Natural Science Foundation of China (No. 61806047). 
We thank Miss Zixian Li and Mr. Guoxian Li for their important discussion.

\section*{Conflict of Interest}
The authors declare that they have no conflict of interest.

\bibliographystyle{elsarticle-num} 
\small
\bibliography{Wanli}

\end{document}